\documentclass[10pt]{article}
\usepackage{a4wide}
\usepackage{graphicx}
\usepackage{float}
\usepackage{amssymb,amsmath} 

\makeatletter

\newcommand{\keywords}[1]{
\subsection*{Keywords}
#1}

\newcommand{\terms}[1]{
\subsection*{General Terms}
#1}

\newcount\catcount
\global\catcount=1

\def\category#1#2#3{%
\ifnum\catcount=1
\subsection*{Categories and Subject Descriptors}
\advance\catcount by 1\else{\unskip; }\fi
    \@ifnextchar [{\@category{#1}{#2}{#3}}{\@category{#1}{#2}{#3}[]}%
}

\def\@category#1#2#3[#4]{%
    \begingroup
        \let\and\relax
            #1 [\textbf{#2}]%
            \if!#4!%
                \if!#3!\else : #3\fi
            \else
                :\space
                \if!#3!\else #3\kern\z@---\hskip\z@\fi
                \textit{#4}%
            \fi
    \endgroup
}

\makeatother
\date{}

\title{{\bf Multiple Retrieval Models and Regression Models \\ for Prior Art Search}}
\author{{\normalsize Patrice Lopez$^{*}$ and Laurent Romary$^{\dag}$ }\\
        {\normalsize Humboldt Universit\"at zu Berlin - Institut f\"ur Deutsche Sprache und Linguistik} \\
        {\normalsize $^{\dag}$INRIA - Gemo Research Group} \\
        {\normalsize $^{*}$European Patent Office, Berlin} \\
        {\normalsize {\tt patrice\_lopez@hotmail.com laurent.romary@loria.fr}}}
\date{}

\begin{document}
\maketitle
\thispagestyle{empty}
\begin{abstract}

\normalsize\noindent 
This paper presents the system called PATATRAS (PATent and Article Tracking, Retrieval and AnalysiS) realized at the Humboldt University for the IP track of CLEF 2009. Our approach presents three main characteristics:
\begin{enumerate}
\item The usage of multiple retrieval models (KL, Okapi) and term index definitions (lemma, phrase, concept) for the three languages considered in the present track (English, French, German) producing ten different sets of ranked results.\item The merging of the different results based on multiple regression models using 
an additional validation set created from the patent collection.
\item The exploitation of patent metadata and of the citation structures for creating restricted initial working sets of patents and for producing a final re-ranking regression model. 
\end{enumerate}
As we  exploit specific metadata of the patent documents and the citation relations only at the creation of initial working sets and during the final post ranking step, our architecture remains generic and easy to extend.   

\end{abstract}




% % %

\section{Motivation}

\noindent 

Our participation to the CLEF IP track was first motivated by our interest in infrastructures for technical and scientific literature in general. 
A large  collection of patent publications offers an excellent opportunity of experimentation. With several millions of documents, such a collection first corresponds to a realistic volume of documents comparable with the largest existing article repositories. Patents cover multiple technical and scientific domains while providing rich cross-disciplinary relations. This level of complexity regarding multiple thematics is similar, for instance, to a large scale repository such as HAL (Hyper Article en Ligne) \cite{rom:arm}. In addition, the European Patent (EP) publications present a quite unique multilingual dimension, often combining three languages (English, French and German) in the same publication (title and claims). Finally, patents can be qualified as extreme exemples of noisy, deliberatly vague and misleading wordings for the title, abstract and claims parts while maintaining relatively standard technical terminologies in the description bodies. 

Our second motivation was to experiment a few fundamental approaches that we consider central and constant for any technical and scientific collections, namely first the exploitation of rich terminological information and natural language processing techniques, second the exploitation of the relations among citations and, third, the exploitation of machine learning for improving retrieval and classification results. Obviously none of these points is original, but we believe that their appropriate combination can provide a framework that would provide much more than the sum of these individual parts. 

Last, we consider that the efficient exploitation and dissemination of patent information are currently not satisfactory. 
While such services as Google Patent and SumoBrain have certainly improved this aspect, patent information is still very difficult to access and exploit as technical documentation. The applicants have usually no interest to disclose their invention in a manner that could facilitate its dissimination. The patent document itself is often poorly structured and follow only a minimal review during its examination focused on the claims and legal aspects. 
However, a clear disclosure of an invention is the counterpart toward the public of patent monopoly rights. 
A patent that is impossible to retrieve is in practice a failure of the patent "contract".  
This is not desirable 
because it is directed against the public and against the goal of a patent system, which is first to motivate and encourage research and innovation. 
Economic and commercial incentives are not the only factor for boosting invention and innovation, 
openly exchanged knowledge is the source and the lifeblood of research and new ideas. For this purpose, better tools for searching and discovering technical and scientific information are also desirable for patent information, beyond pure economic aspects.

\section{CLEF IP 2009 and the Prior Art Task}

Before starting to describe our system, we introduce some basic definitions and examine in more detail the task and how it defers from a standard prior art search for patent examination. We also discuss the main differences between a patent document and traditional documents considered in usual text retrieval tasks. 

In the following description, the "collection" refers to the data collection of approx. 1,9 millions documents corresponding to 1 million European Patents. This collection represents the prior art. The "training set" refers to the 500 documents of "training topics" provided with judgements (the relevant patents to be retrieved). The "validation set" corresponds to a subset of approx. 4000 patents selected by us from the collection and a "patent topic" refers to the patent for which the prior art search is done. 

\subsection{Prior Art Searches}

The goal of the CLEF IP 2009 track was to identify in the collection the closest prior art to a given patent. The evaluation was produced automatically using patent citations introduced during the official prosecution of this patent application at the European Patent Office (EPO). The list of patent citations, therefore, gathers the patent citations provided by the applicant himself, the result of the prior art search performed by the patent examiner, and patent citations introduced at a later stage of proceedings (examination and possibly opposition). 

The prior art search as implemented in the CLEF IP 2009 track can be considered globally as easier than a real one. The usual starting point of a patent examiner performing a prior art task is a set of application documents in only one language, with one or more IPC\footnote{International Patent Classification: a hierarchical classification of approx. 60.000 subdivisions used by all patent offices and maintained by the WIPO (World Intellectual Property Organization). }  classes and with a very broad set of claims. 
In the present task, the topic patents were entirely made of examined granted patents providing reliable information resulting normally from the search phase: 
\begin{itemize}
\item The ECLA\footnote{European CLAssification: an extension of the IPC corresponding to a hierarchical classification of approx. 135 600 subdivisions, about 66 000 more than the IPC. } classes. They corresponds to a fine classification used for the search, more precise than the IPC. 
\item The claims of the granted patents. These final claims are drafted taking into account the prior art identified in the search report and are often revised for removing clarity issues. In addition, this final version of the claims is translated in the three official languages of the EPO by a skilled human translator, making crosslingual IR techniques more reliable.
\item A revised description. The first part of the description often acknowledges the most important document of the prior art which has been identified during the search phase.
\end{itemize}

While such information alone is not sufficient to perform reliable prior art analysis, it can be combined to more standard text retrieval techniques in order to achieve higher accuracy.

It is also important to stress that other aspects make the task more difficult than a prior art search based on a fresh patent application.
An examiner can exploit some very useful information as the patent families which permit to relate patents from different patent systems (PCT, US, etc.). The examiner can also search non EP publications and get search report from other patent offices in case the application has already been searched elsewhere. 
Moreover, as the evaluation is made on the basis of the patents cited in the EPO search reports and cited during examination and possible opposition phases, the list of relevant documents is by nature partial and often motivated by procedural purposes. Typically an examiner stops his research when he has found a very good X document that will be used to refuse an application. The goal of a patent examiner is not really to search the best of all relevant documents, but a subset or a combination that will support his argumentation during the examination phase. 
Very good prior art documents that are difficult to find will often not appear in the search report and a very good automatic retrieval system that could present this document highly ranked will not been rewarded.  In addition, as non patent literature is not considered in this task, the collection and the evaluation can be viewed as biaised: when an examiner finds one or two good non patent documents, he may limit his further search in the patent database to complementary documents covering minor embodiment aspects. It must be noted that the non-patent documents (mainly journal and conference articles) represent at the EPO more than 30\% of the citations in several technical domains, for instance more than 60\% in biotechnologies. These different aspects make the final results relatively difficult to generalize to a standard prior art search.

The task remains, however, a good approximation and we consider that all the techniques presented in this work remain pertinent for standard prior art searches. In addition, these conditions fit well an "invalidity search" made by a third party after the grant of patent for engaging a possible opposition. Finally, the task is also relevant for the purpose of a technical survey, from the point of view of the scientist or the engineer willing to evaluate the novelty and inventiveness of a work in term of the patent law.  

\subsection{The limits of the textual content}

The textual content of patent documents is known to be difficult to process with traditional text processing and text retrieval techniques. As pointed out by \cite{kri:zac}, patents often make use of non standard terminology, vague terms and legalistic language. The claims are usually written in a very different style than the description. The description also frequently contains digressions and general presentations of the technical fields which do not provide any useful information about the contribution of the patent. 
A patent also contains non-linguistic material that could be important: tables, mathematical and chemical formulas, citations, technical drawing, etc.
For so called drawing-oriented fields (such as mecanics), examiners focus their first attention only on drawings and we can suppose that any automatic retrieval based only on text will fail. 

So one could challenge the relevance of any standard technical vocabulary for searching patent documents. However, the description, after a general presentation of the state of the art, illustrates the claimed "invention" with preferred embodiments which very often use a well accepted technical terminology, and exhibit a language much more similar to usual scientific and technical literature. 

\subsection{The citation structures}

The patent collection is a very dense network of citations creating a set of interrelations particularly interesting to exploit during a prior art search. Table \ref{citation} gives a quantitative overview of the citation network we observed for the patent collection\footnote{Just as a comparison, following Thomson Reuters' Journal Citation Reports, approximately 60 millions citations has been made in more than 7000 journals during the time period 1997-2005. }. The large majority of patents are continuations of previous works and previous patents. The citation relations make this development process visible. Similarly, fundamental patents which open new technologies subfields are exceptional but tends to be cited very frequently in the whole subfield during years. 

\cite{clark}, for instance, exploits the citation graph of a patent collection for identifying patent thickets, i.e. the patent portfolios of several companies overlapping on a similar technical aspect. While the author's goal was to identify pro-competitive technical domains with respect to antitrust regimes, this work shows that multiply related patents can be infered from the overall citation network of a patent collection. High density of inter-citations between a group of applicants is the evidence of cumulative innovation and multiple blocking patents. If a new patent applicant belonging to this patent ticket appears, it is very likely that the most relevant prior art documents are already present in this patent thicket. 

In addition, as drawings are excluded from the present task, the exploitation of citation relations appears the best source of evidence for retrieval in drawing-oriented fields. 

\begin{table}[ht]
\centering
\begin{minipage}{\textwidth}
\centering
\begin{tabular}{ | l | l | r | }
\hline
Citations & Category\footnote{X means that the cited document taken alone anticipates the claimed invention. Y indicates that the cited document in combination with other prior art documents covers the claimed invention. A means that the document is relevant but discloses only partially the claimed invention. Finally, D, which can appear together with the previous categories, indicates that the document has been cited by the applicant in the original description.} & \# \\
\hline
total & all  & 4 854 280\\  
& X &  581 853 \\ 
& Y & 413 981 \\ 
& A & 1 849 251 \\ 
& D & 2 019 733 \\ 
& other & 198 749 \\
\hline
EP doc. & all & 1 082 647\\
EP doc. with citation text & all & 363 494\\
\hline

\end{tabular}
\end{minipage}
\caption{\small Overview of citation relations in the patent collection. }
\label{citation}
\end{table}

The patents cited in the description of a patent document are potentially highly relevant documents. First, the examiner often acknowledges the applicant's proposed prior art by adding this document in the search report (usually as an A document since it is extremely rare that an applicant discloses himself, by mistake, a "killing" X document in his application). Second, in case the patent document corresponds to a granted patent, Rule 42(1)(b) EPC requires the applicant to acknowledge the closest prior art. As a consequence, the closest prior art document, sometimes an EP document, is frequently present in the description body of a B patent publication.   

Actually, we observed that, in the final XL evaluation set, the European Patents cited in the descriptions represent 8,52\% of the expected prior art documents of the final topic set. For the 10.000 topic patents, we found that 4407 were citing in their descriptions at least one EP document from the collection, for a total of 7960 cited EP documents for which 5305 are relevant. This is 21.66\% of the relevant documents of these 4 407 topics. With a "run" made only with these cited patents, a MAP (Mean Average Precision) of 0.2230 is obtained, with a precision at 5 of 0.2315. This shows the potential contribution of this simple source of relevant patents, while it must be noted that it is only a partial source since it leaves more than half of the patent topics without any citations.  

\subsection{Importance of metadata}

In addition to the application content (text and figures) and the citation information, all patent publications contain a relatively rich set of well defined metadata. 
Traditionally at the EPO, the basic approach to cope with the volume of the data is, first, to exploit the European patent classification (ECLA classes) to create restricted search sets and, second, to limit the broad searches to the titles and abstracts. When following this strategy, an examiner has retrieved a set of approx. 50 documents, he visualizes each of these documents and performs a careful non automated analysis of the whole content including the description text and the figures. Exploiting the ECLA classes appears, therefore, as a solid basis for efficiently pruning the search space. 

In addition to the classification metadata, during 30 years of prior art search, the EPO patent examiners have developed heuristics for finding interesting documents given the application in hand. Since the means for searching text content are still relatively rudimentary (in 2009, only a command line boolean search engine is used at the EPO), these heuristics are often based on metadata such as the applicant name, the inventor names, other patent office classes or priority documents. 

\subsection{Multilinguality}

The European Patent documents are by definition highly multilingual. First, each patent is associated to one of the three official languages of application. In the collection, the language distribution was the following: 69\% for English, 23\% for German and 7\% for French. It indicates that all the textual content of a patent will be available at least in this language. Second, granted patents also contains the title and the claims in the three official languages. These translations are usually of very high quality because they are made by professional human translators. Crosslingual retrieval techniques are therefore crucial for patents, not only because the target documents are in different languages, but also because a patent document often provides itself reliable multilingual information which makes possible  the creation of valid queries in each language.

As it is important not to limit a retrieval only in the main language of the patent application, our system needed to deal with different languages for each patent. We have thus decided to build different index and different specialized retrieval models for each languages and, in a second time, to merge the different results in oder to exploit the benefit of each language. 

\section{Overall Description of the System}

\subsection{System architecture}

\begin{figure}[H]
  \begin{center}
    \includegraphics[width=15.5cm]{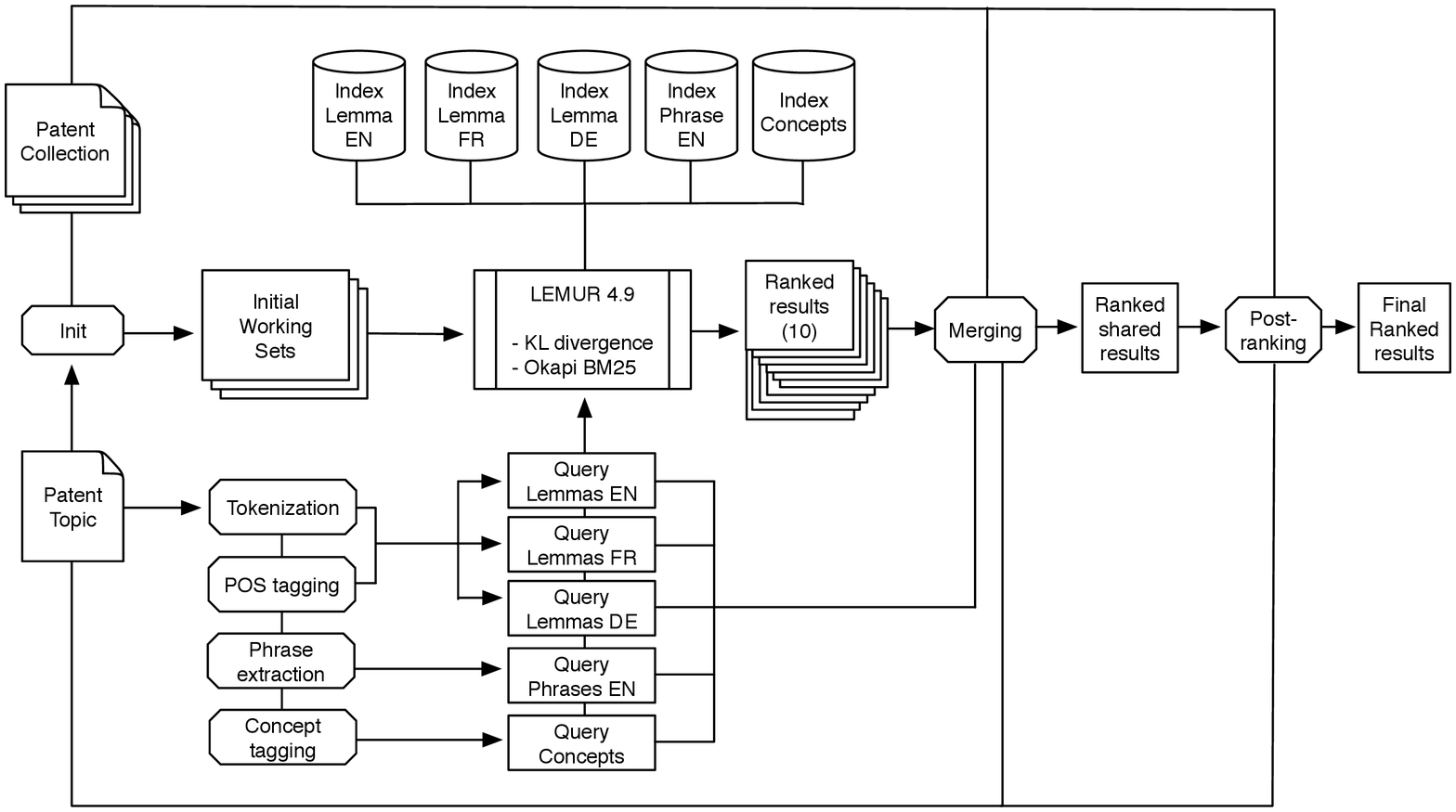}
  \end{center}

  \caption{\small System architecture overview of PATATRAS for query processing. The arrow represents the main data flow from the patent topic to the final set of ranked results.}
  \label{system}
\end{figure}

As explained in the previous section, there is clear evidence that pure text retrieval techniques are insufficient for coping correctly with patent documents. Our proposal is to combine useful information from the citation structure and the patent metadata, in particular patent classification information, as pre and post processing steps of text-based retrieval techniques. In order to exploit multilinguality and different retrieval approaches, we merged the ranked results of multiple retrieval models based on machine learning techniques. As illustrated by Figure \ref{system}, our system, called PATATRAS (PATent and Article Tracking, Retrieval and AnalysiS), relies on four main processing steps:
\begin{enumerate}
\item the creation of an initial working set for each patent topic in order to limit the search space,
\item the application of multiple retrieval models (KL divergence, Okapi) using different index models (English lemma, French lemma, German lemma, English phrases and concepts) for producing several sets of ranked results,
\item the merging of the different ranked results based on multiple SVM regression models and a linear combination of the normalized ranking scores,
\item a post-ranking based on a SVM regression model.
\end{enumerate}

Steps 2 and 3 have been designed as generic processing steps that could be reused for any technical and scientific content, patent or non patent. The step 2 uses standard text retrieval techniques. The step 3 uses domains information and standard metadata such as the language that we can find or obtain automatically in any type of document collections. The patent-specific information are exploited in steps 1 and 4. Before presenting in more in details these four different steps, we now briefly describe how all the patent documents have been parsed and pre-processed.

\subsection{Document parsing}

%\noindent
For each patent in the collection, training and topic set, the following processing were performed:
\begin{enumerate}
\item Parsing of the initial XML documents corresponding to the patent (so called {\sl A} document before a decision and {\sl B} documents in case of grant).
\item Representation of all the metadata in a MySQL database, following a comprehensive relation model - given the required heavy processing based on the metadata, and although XML databases can present some advantages,  we needed a very fast and easy to optimize database.  
\item Identification by means of regular expressions of the patents cited in the latest version of the description (so from a decreasing order of priority: B2, B1, A2, A1).
\item For all the textual data associated to the patent: Rule-based tokenization depending on the language.
\item For all the textual data associated to the patent: Part of speech tagging and lemmatization. We used our own HMM-based implementation for English and  the Tree Tagger \cite{hel:sch} for French and German.
\end{enumerate}

\subsection{Metadata Database}

We performed a basic normalization and cleaning of all inventor and applicant names: Particules and titles were removed from the inventor fields (for instance {\sl Professor Dr. Dr. h.c. mult. Wolfgang  Wahlster} becomes simply -with all due respect- {\sl Wolfgang Wahlster}), business entity marks (Inc., GmbH, Kabushi Kaisha) and locations (country names) were removed from applicant fields. 
We also stored the citation texts of the patents cited in a description. The database storing all metadata of the collection and all corresponding indexes, but not the textual content, had a final size of 2,48 GB.

\section{Indexing models}

\subsection{Overview}
The five following indexes were build using the Lemur toolkit \cite{lemur} (version 4.9):
\begin{itemize}
\item For each of the three language (English, French, German), we built a full index at the lemma level. 
\item For English, we created an additional phrase index based on phrase as term definition.
\item A crosslingual concept index was finally built using the list of concepts identified in the textual material for all three languages. 
\end{itemize}

We did not, therefore, index the collection document by document, but rather considered a "meta-document" corresponding to all the publications related to a patent application. In each case, the following textual data corresponding to a patent is indexed:
\begin{itemize}
\item the last version of the title,
\item the first version of the description (Following Article 123(2) EPC, we are sure that any further publication will not go beyond the scope of the initial version of the description),
\item the last version of the abstract,
\item the last version of the claims.
\end{itemize}

\subsection{Lemma indexes}

Based on the result of the lemmatization, we considered the lemma present in the textual data of the publications corresponding to the patents of the collection: title, abstract, claims of the last available publication, and description of the earliest available publication - so A1, A2 or B1 in this order. The selection of the lemma as term unit could be view as a stemming removing all inflexional suffixes. Only the lemmas corresponding to open grammatical categories (i.e. noun, verb, adverb, adjective and number) has been indexed, which could be viewed as applying a stop-word list in traditional IR.

\subsection{Phrase indexes}

For the English content, a phrase extraction was realized: based on the part of speech tagging results, all the noun phrases were identified and the Dice Coefficient was applied to select the phrases \cite{smadja}. 

\subsection{Multilingual Terminological database}

%\noindent
A multilingual terminological database covering multiple technical and scientific domains and based on a conceptual model, have been created from the following existing "free" resources: MeSH, UMLS, the Gene Ontology, a subset of WordNet corresponding to the technical domains as identified in WordNet Domains \cite{mag:cav} with the corresponding entries of WOLF (a free French WordNet), and a subset of  the English, French and German Wikipedia corresponding to technical and scientific categories \cite{wiki}. 

The Wikipedia "dump" XML files were processed with a slightly modified version of Wikiprep \cite{wikiprep} able to extract multilingual relations in addition to usual structure and text information. Similarly as in \cite{gab:mar}, we interpreted an article as a concept, the title of an article being the preferred term and the disambiguation redirections to this article being alternative or variant terms realizing this concept.

\begin{figure}[!b]
  \begin{center}
    \fbox{\includegraphics[width=15cm]{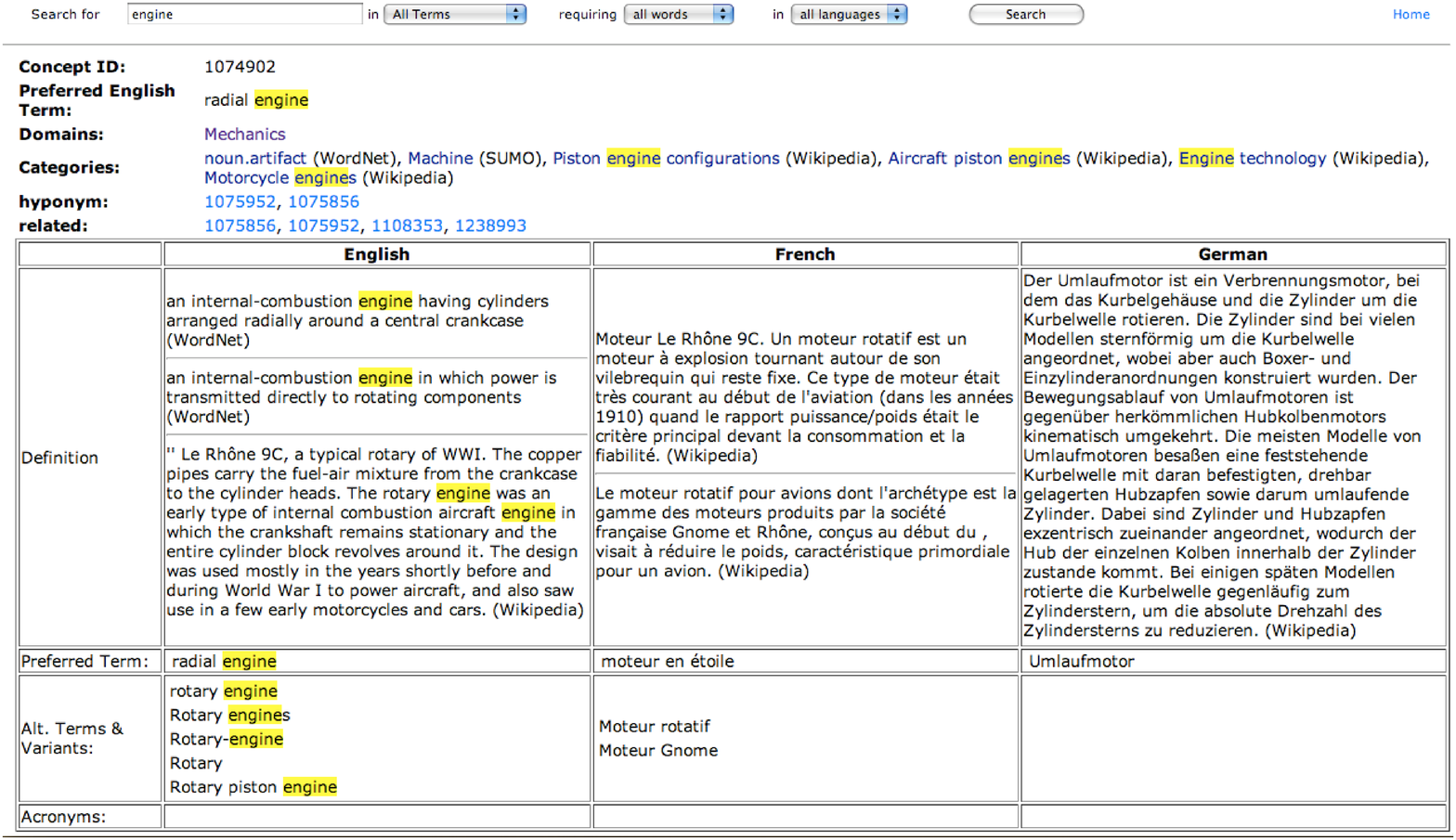}}
  \end{center}

  \caption{\small View of the multilingual terminological database for the  concept corresponding to the term {\sl Radial Engine}.}
  \label{grisp}
\end{figure}

A set of 76 general technical domains have been first established derived from the Dewey Decimal Classification \cite{dewey}. For each above-mentioned sources, a mapping has been written between the main catgories/domains of the source and the general domains. Two concepts coming from two different sources were merged if they share at least one term and one domain.

The resulting database contains:
\begin{itemize}
\item Approx. 3 millions terms (2,6 millions for English, 190.000 for German, 140.000 for French),
\item 1,4 millions of concepts (71.000 realized in German and 65.000 realized in French),
\item 600.000 definitions,
\item 1 million of semantic relations,
\item approx. 20.000 fully specified acronyms,
\item 123.000 additional "source-specific" categories.
\end{itemize}

The terminological database relies on a conceptual model \cite{romary} and is currently implemented in a MySQL database. A web interface has been developed for browsing the terminological database and for performing basic searches, see Figure \ref{grisp}. For the purpose of the CLEF IP tasks, we use only the terms and the acronyms information of this terminological resource. 

Wikipedia offers a massive sources of terminological data (more than 1,5 million of terms for the technical and scientific categories) and was the only multilingual source. However, the quality of the data is clearly not comparable with a well designed domain specific  terminology as for instance MeSH. In particular, due to overall noisy data, the merging of concepts involving Wikipedia entries was not satisfactory. In practice, it was not possible to consider as term variants the disambiguation terms given by Wikipedia and we merged concepts involving Wikipedia entries only if the article-level term entry was in common with the other concepts for a common domain. This work is still ongoing and we expect to improve it in the future. 

\subsection{Concept tagging} 

The terms of the terminological database have been used for annotating the textual data of the whole collection, training and topic sets. A term annotator able to deal with such a large volume of data has been developed specifically for this track. This annotator can match term variants following morphological variations. The concept disambiguation was realized on the basis of the ECLA classes (or by default the IPC classes) of the processed patent. For this purpose, the upper level of the IPC  (corresponding to the IPC "classes" as defined in the IPC, i.e. the three first characters of the classes/subclasses/groups/subgroups) has been mapped to the common abovementioned  76 domains. Given the design of the terminological database, a term used in a given domain corresponds to a single concept. When an IPC class corresponds to several domains (for instance the class G10 is used for music instruments which can correspond to the domains {\sl acoutics} and {\sl electronics}), or when a term corresponds to several concepts in the same domain (for instance {\sl engine} in the computing field which could be a {\sl layout engine} or a {\sl search engine}), a term cannot always be disambiguated on the basis of the IPC class. In this case, we have decided in the present implementation not to further attempt to select a concept and to skip this term.  

\subsection{Retrieval models}

We used the two following well known retrieval models: 
\begin{itemize}
\item Unigram language model with KL-Divergence and Jelinek-Mercer smoothing ($\lambda=0.4$),
\item Okapi weighting function BM25 ($K1=1.5, b=1.5,K3=3$).
\end{itemize}
The two models have been used with each of the previous five indexes, resulting in the production of 10 lists of retrieval results for each topic patent.

In each case the query was build based on all the available textual data of a topic patent and processed similarly as the whole collection in order to create one query per language based on lemma, one query based on English phrases and one query based on concepts (independent by definition from the language). The query for a given model is, therefore, a representation of the whole textual content of the patent. The query is exactly the same representation as the one of a document indexed in the collection. The scoring model used for retrieval corresponds, in this context, to a distance between two "documents". In the framework of language model based IR, KL divergence is typically used for evaluating the distance between two documents, i.e. between two unigram probabilistic distributions. While Okapi BM25 is usually used for ad hoc retrieval, it is also known as a reliable scoring function for evaluating the distance between documents. The drawback of using a whole document representation as query is the processing time which is always related to the size of the query. However, for the present work, we did not consider processing time as an issue, as long as the whole set of patent topics could be processed in the track timeframe.  
For both retrieval methods, we did not use query expansion, nor pseudo-relevance feedback, because these two techniques did not appear effective during our first experiments. 

The retrieval processes were based on the Lemur toolkit \cite{lemur}, version 4.9.
The baseline results of the different indexes and retrieval models are presented in Table \ref{models}, column (1). These results correspond to the application of the retrieval model with on whole collection. 

\begin{table}[ht]
\centering
\begin{tabular}{ l l l c c c c }
\hline
Model 
& 
Index 
& 
Language 
& 
(1)
& 
(2)
& 
(3)
& 
(4)\\
\hline
KL & lemma & EN & {\bf{0.1068}} & {\bf{0.1083}} & {\bf{0.1516}} & {\bf{0,1589}} \\   
KL & lemma & FR & 0.0611 & 0.0612 & 0.1159 & 0,1234 \\  
KL & lemma & DE & 0.0627 & 0.0634 & 0.1145 & 0,1218 \\  
KL & phrase & EN & 0.0717  &  0.0720 & 0.1268 & 0,1344\\   
KL & concept & all & 0.0671 & 0.0680 & 0.1414 & 0,1476 \\   
Okapi & lemma & EN & 0.0806 & 0.0813 & 0.1365 & 0,1454\\  
Okapi & lemma & FR & 0.0301 & 0.0303 & 0.1000 & 0,1098 \\  
Okapi & lemma & DE & 0.0598 & 0.0612 & 0.1195 & 0,1261 \\  
Okapi & phrase & EN & 0.0328 & 0.0330 & 0.1059 & 0,1080\\ 
Okapi & concept & all & 0.0510 & 0.0516 & 0.1323 & 0.1385 \\ 
\hline

\end{tabular}
\caption{\small MAP results of the retrieval models, Main Task.  (1) Base MAP, Medium set (1 000 queries), (2) MAP with citation texts, Medium set, (3) MAP with citation texts, initial working sets, Medium set, (4) Map with citation texts, initial working sets, XL set (10 000 queries).}
\label{models}
\end{table}

We can observe that for the same query and the same index, KL divergence always provided better results than Okapi. The best result is obtained with KL divergence with English lemma representation. The fact that conceptual and phrase based retrievals perform worst than the monolingual English lemma representation can appear disappointing given the effort needed to implement them. However, it is consistent with previous works which have noted the information loss implied by pure conceptual representations as compared to simple stem-based retrieval. 
 
\subsection{Citation texts}

The citation texts of a target patent are all the paragraphs in other patent documents that refers to it. Extending the content of a patent with its citation text aims at providing more textual descriptions corresponding to the important contributions of the patent according to the other patent applicants. Following a similar approach, \cite{nak:sch}, for instance, tried to exploit the citation texts in order to improve the semantic interpretation and the retrieval of text for biomedical articles. Moreover, for the case of patent documents, since part of the citation text can be written in the other languages than the taget patent, it possibly increases the multilingual description of the cited patents. While for technical and scientific articles, the citation text is usually just a sentence, citation texts for patents appear to be in a constant manner a whole paragraph. Therefore, for each patent document present and cited in the collection, the entire paragraph of citation was appended to the textual material of the cited patent.  

The table \ref{models} presents the impact of adding the citation text, see column (2). The improvement is low and statistically not significant. We think, however, that this result is encouraging because it was obtained with a very limited number of citation texts. Only citations of an EP document were considered here. By having a complete collection or patent family information, it would be possible to extract much more citation texts (most likely by a factor five to ten) and we could expect that the improvement would be more significant. 

\section{Creation of initial working set}

For each topic patent, we created a prior working set for reducing the search space and the effect of term polysemy. The goal here is, for a given topic patent, to select the smallest set of patents which has the best chance to contain all the relevant documents. A set $S_p$ of patents for a given patent $p$ is created by applying successively the following steps:
\begin{enumerate}
\item Put in $S_p$ all the patents cited in the description of the topic patent (as identified in step 3 of the document parsing) and present in the collection.
\item Up the citation tree: All patents citing at least one patent of $S_p$ are added to $S_p$.
\item Down the citation tree: All patents cited by at least one patent of $S_p$ are added to $S_p$
(steps 2 and 3 were performed iteratively a second times after step the fifth step).
\item Priority dependencies: All patents having a priority document in common with $p$ or with a patent in $S_p$ are added in $S_p$. All patents citing a priority document of a patent in $S_p$ are added to $S_p$. This step permits to gather non unitary and divisional patent applications (i.e. a single patent application containing possibly different independent inventions which results in several parallel applications) and to exploit the partial  information present in the collection about patent families.
\item All patents having the same applicant as the topic patent and at least one common inventor are added in $S_p$.
\item All patents belonging to one of the ECLA class of $p$ (if at least one ECLA class is available) are added in $S_p$.
\item All patents belonging to the ECLA classes most frequently co-occurring in the collection with the ECLA classes of p are added in $S$.
\item If the working set is below a given limit: All patents having the same applicant and belonging to one of the IPC class of $p$ are added to  to $S_p$.
\item If the working set is still below a given limit: All patents belonging to one of the IPC class of $p$ are added to  to $S_p$.
\end{enumerate}

\begin{table}[ht]
\centering
\begin{tabular}{l r l r}
\hline
Step & \# relevant doc. & micro recall &\# docs \\
\hline
1. cited patents & 5 305  & 0.0852  & 7 960 \\
2. up the citation tree &  5 853   & 0.0940 (+10,3\%) &  20 245 \\
3. down the citation tree & 6 026  & 0.0967 (+2,9\%)   & 22 727 \\
4. common prority doc.: & 7 999 & 0.1284 (+32,78\%) & 42 168 \\
5. same applicant/inventor & 10 768  & 0.1728 (+34,58\%)  & 116 099 \\
2+3. second iteration &  11 463  &  0.1840 (+6,48\%)  & 176 815\\
6. same ECLAs & 35 935  & 0.5769 (+213,53\%) & 5 129 068 \\ 
7. most freq. co-occuring ECLA & 44 559  & 0.7154 (+24,22\%)  & 24 859 957 \\ 

8. same applicant and IPC & 44 625  & 0.7164 (+0.14\%) & 24 863 768\\

9. same ipc & 45 489 & {\bf{0.7303}} (+1.94\%)  & 26 159 190\\  
\hline

Total relevant & 62 285 & 1.0 & \\
\hline
\end{tabular}
\caption{\small Incremental construction of the initial working lists of patents for the Main XL task. }
\label{working}
\end{table}

At the end of this process, if the size of the working set is too low or too large, no working set is used and the retrieval is performed on the whole collection. In the submitted runs, the lower limit was 10 documents and the upper limit 10.000. The table \ref{working} presents the performance of each step in term of increase of the micro recall, i.e. coverage of all the relevant documents of the whole set of topics. Note that the recall reported in the official CLEF IP 2009 evaluation summary is the macro recall, i.e. the average of recall obtained for each topic patent. 
The last column gives the sum of the number of documents for all working sets.

The micro recall of the final run was 0.6985, thus lower than the one of the final working lists. This difference comes from the working list having a number of patents higher than 1000 and where the final processing has not been able to place all the relevant documents in the first 1000 patent results. These "missed" 
patents are the most difficult documents to process: the system failed to rank them both on the basis of the textual content and the metadata even with an initial working set. Since many working lists contained less than 1000 documents and sometimes less than 100 documents, the list of results of the final run was frequently less than 1000. Similarly some working sets were particularly large, sometimes more than 10 000 documents, but all final runs were cut at 1000. The final number of results was, therefore, on average approx. 415 documents per patent topic, as compared to approx. 2616 documents in average per initial working set. 

It is also clear that we are missing 26.97\% of the expected relevant documents which is a relatively high number. Identifying how to capture these documents without expanding too much the working sets will require further investigations. 

These different steps correspond to typical search strategies used by the patent examiners themselves for building sources of interesting patents. They capture techniques that have emerged in patent examination and which are considered to be effective. As the goal of the track is, to a large extend, to recreate the patent examiner's search reports, recreating such restricted working sets appears to be a valuable approach.
Table \ref{models}, column (3) and (4) show the improvement of using the initial working sets instead of the whole collection. The retrieval with working sets was realized using the "working list" functionality of Lemur in batch mode.

\section{Merging of results}

In the previous section, we have observed that conceptual and phrase-based retrievals alone were less efficient than English lemma model. However, several works have shown that the conceptual and phrase representation can improve a word-based model, for instance the semantic smoothing techniques in the language model framework. In our preliminary results, we observed that the different retrieval models present a strong potential of complementarity, in particular between the different languages and between lemma/concept. The table \ref{complementarity} presents the repartition of the best results over the different models. We can observe that concept-based models provide a high number of results with higher MAP than the English lemma model. We can also note that each language-specific model provides a constant set of best results over all other models. Intuitively,  for a topic patent essentially described in the main  language of application (in particular the whole description) and minimally in the two other languages (just the title and the claims), the index corresponding to main language should provide better results and should be prioritized.  

\begin{table}[ht]
\centering
\begin{tabular}{l c c r}
\hline
Model &  \begin{tabular}{c} \# better than \\ baseline \\ \end{tabular} & \# equal & \begin{tabular}{r}  \# best \\ overall \\ \end{tabular} \\
\hline
\begin{tabular}{l}
KL lemma en  \\
(baseline) \\
\end{tabular} & -  & - &   1341 \\
KL lemma fr & 3480 & 661 & 839 \\
KL lemma de  & 3392& 632 & 781  \\
KL phrase en  & 3559& 741 & 869 \\
KL concept  & {\bf 4832} & 133 &  {\bf 1692} \\
Okapi lemma en & 3928 & 789 & 956 \\
Okapi lemma fr & 3224 & 616 & 626 \\
Okapi lemma de  & 3494 & 630 & 836 \\
Okapi phrase en  & 3002 & 649 & 572 \\
Okapi concept  & 4638 & {\bf 114} & 1488 \\
\hline
\end{tabular}
\caption{\small Complementarity between results sets for the XL patent topic set (10 000 documents). }
\label{complementarity}
\end{table}

Merging multiple results has been used in the context of distributed information retrieval, in particular with partially overlapping collections. 
Merging ranked results from different models for the same collection appears well adapted to a patent collection because the different models exploit different views, for instance different languages, for retrieving documents in the same collection. 
In addition, we are presently in an exceptional situation where we can exploit a very large amount of training data given  
the number of citations present in the collection. 
For training purposes, 500 complete examples were provided with judgements. Moreover, the collection itself could provide a huge number of examples of prior art results. 
This uncommon aspect makes possible a fully supervised learning method. 
Machine learning algorithms are, therefore, well appropriated to weight the different ranked lists so that, given a query, the most reliable models are prioritized. The merging of ranked lists of results is here expressed as a regression problem.

The usage of regression models for merging results  was described for instance in \cite{savoy} and \cite{si:cal}.  Merging based on regression usually surpasses other combination methods which do not involve machine learning, such as the well known CORI merging algorithm. In \cite{si:cal}, the precision following a merging of results from different search engines was improved up to 98.9 \% as compared to a merging based on the CORI algorithm. In \cite{savoy}, even with a very limited number of features used for learning, a merging of retrieval results from different languages based on a logistic regression significantly surpasses all other score combination approaches. Regression models appears particularly appropriate in the present case, because they  permit to adapt the merging on a query-by-query basis.

For each model $m$, we trained a linear regression model using as input a set of features inferred from the query. As a general framework, given a set of examples,  
$(\overrightarrow{x_1} , y_1 ), . . . , (\overrightarrow{x_N}, y_N)$, where the $\overrightarrow{x_i}$ are vectors of features and the $y_i$ are values corresponding to the dependent variable, the goal of a linear regression is to find $\overrightarrow{w}$ such that 
and 
\begin{equation}
\Phi(\overrightarrow{w}) = min  \displaystyle\sum_{i}(y_i-\overrightarrow{w}.\overrightarrow{xi})^2
\end{equation}
In the present regression training, the observed MAP was used as the dependent variable for representing the pertinence of a set of results for a given retrieval model. 

For the realizing the merging, scores were first normalized so that they all lie in the same range. We used the standard normalization, basically all the min are shift to 0 and the max is scaled to 1: 
\begin{equation}
w_{i} = \frac{v_{i} - v_{min}} { v_{max} -  v_{min} }
\end{equation}
The regression model gives a  score $c_{qm}$ for the retrieval model $m$ and for the query $q$ which is interpreted as an estimation of the relevance of the results retrieved by $m$ for the query $q$. 

The merged relevance score  $s_{dq}$ for a patent $d$ following a query (here a patent topic) $q$ is obtained as a linear combination of the normalized scores $w_{md}$ obtained from each retrieval model $m$:
\begin{equation}
s_{dq} = \displaystyle\sum_{m \in M} c_{mq} w_{md} 
\end{equation}
This score is computed for each patent appearing in at least one of the result sets produced by the different models. The merged result set is build by ranking these patents according to this new relevance score.

The key of the supervised learning approach is to exploit as much training data as possible in order to use a rich set of features while avoiding overfitting. To exploit more training data, we created from the collection track a supplementary training set of 4 131 patents.  We selected patents citing at least 4 EP documents, with a language distribution and an IPC class distribution similar to the ones of the whole collection. We assembled the initial working sets similarly as explained in section 5, but we filtered out patents whose publication dates were after the priority date of the considered patent. During our tests for selecting the regression algorithm, we built the merging models based on these 4 131 patents, that we called the validation set, and tested them on the "normal" training set of 500 topics. For producing the final official runs, we trained the merging models on the whole 4.631 training patents.

The following set of features was used: (f1) the language of proceeding of the patent topic, (f2) the size of the query, (f3) the size of the initial working set, (f4-5) the non-normalized minimum and the maximum retrieval scores of the set of results, (f6) the range of the non-normalized score of the result set, (f7-8) the main IPC trunk (first character of an IPC class) and the IPC class (three first characters of an IPC class/group). The average number of words of the phrases was also used for the results based on the phrase index (f9). 

Expressing the merging of ranked results as a regression problem permits to use existing machine learning software packages which provide excellent evaluation utilities. We have experimented several regression models: least median squared linear regression, SVM regression (SMO and $\nu$-SVM methods) and multilayer perceptron using LibSVM \cite{libsvm} for the $\nu$-SVM regression method and the WEKA toolkit \cite{weka} for the other methods. For the $\nu$-SVM regression method, we used the methodology described in \cite{guide} for setting the parameters which includes  scaling and cross-validation. SVM in general is known to be sensitive to hyperparameter selection ($\lambda$). This methodology led to a strong improvement as compared to our first random testing and the SMO regression.

\begin{table}[ht]
\centering
\begin{tabular}{l l l l l l }
\hline
Features & LeastMedSq & \begin{tabular}{c} multilayer \\ perceptron \\ \end{tabular} & SMO  & $\nu$-SVM  \\
\hline
f1 & 0.1681 (+5.8\%) & {\bf{0.1711}} (+7.7\%) &  0.1706 (7.4) & 0.1691 (+6.4\%)\\
f1-6 & 0.1689 (+6.3\%) & 0.1797 (+13.1\%) & 0.1807 (+13.7) & {\bf{0.1976}} (+24.3\%) \\
all  & 0.1786 (+12.4) &  0.1898 (+19.4\%) & 0.2016 (+26.9\%) & {\bf{0.2281}} (+43.5\%) \\
\hline
\end{tabular}
\caption{\small MAP measures following different regression models for merging the 10 ranked results (XL patent topic set, 10 000 documents). The number in parenthesis gives the relative improvement over the best individual score, English lemma KL model at 0,1589. }
\label{combine}
\end{table}

Table \ref{combine} gives the breakdown of our experiments based on the 4631 training patents. The results obtained with the $\nu$-SVM regression model shows that  the methodological aspect is a key for achieving better performances. The default parameters have been used for  the other regression models. The final runs are based on $\nu$-SVM regression employing the Radial Basic Function (RBF) as kernel function. 

We also observed that the IPC class is particularly relevant for identifying strong confidence of the concept retrieval models. The terminological knowledge base contains a very rich amount of terms from the medical, biotech and chemical fields coming from well curated sources (MeSH, UMLS and the Gene Ontology). On the contrary, a field like computer sciences suffers from the noisy Wikipedia data. 

\section{Post-ranking}

Post-ranking, or re-ranking in general, is a particularly attractive technique in Information Retrieval because, first, the usual result of a retrieval model is a weighted n-best list of outputs and, second, it is much easier to experiment and optimize a post-ranker for a particular set of features than to integrate them in a single model. In the present case, as we used retrieval models designed for texts, it would be extremely difficult or impossible to modify these methods for including the features we want, namely to prioritize certain patents based on citation relations and metadata. The drawback of re-ranking is, of course, that it is limited by the initial model.

While the previous section was focusing on {\sl learning to merge ranked results}, this step aims at {\sl learning to rank}. Regression here is used to weight a patent result in a ranked list of patents given a query. The creation of the inital sets was dealing with recall and we address now precision. The goal of the re-ranking of the final merged result is first of all to boost the score of certain patents: 
\begin{itemize}
\item Patents initially cited in the description of the topic patent: a boolean feature was used to indicate if the retrieved patent was present or not in the citations.
\item Patents having several ECLA and IPC classes in common with the topic patent: two features were introduced to indicate the number of common ECLA classes and the number of IPC classes.
\item Patents with higher probability of citation as observed within the same IPC class and within the set of results: we introduced two features corresponding to a number between 0 and 1 representing these two probabilities.
\item Patents having the same applicant and at least one identical inventor: the following features were added: a boolean indicating if the applicant was common between the retrieve patent and the patent topic, and a number between 0 and 1 indicating the proportion of common inventors.
\end{itemize}
Similarly as for the creation of the initial working sets, these features correspond to criteria often considered by patent examiners when defining their search strategies. Re-ranking based on these features permits to obtain a final result closer to a standard EPO search report and thus increases the average precision. 

The dependent variable represents the weight adjustment to be applied to the particular patent result. In the training data, the dependent variable $s_p$ for a patent result $p$ was set as follow: 

\begin{equation}
s_p=\begin{cases} 
w_{max} & \text{if $p$ is relevant},\\ 
0 & \text{Otherwise}. 
\end{cases} 
\end{equation}

where $w_{max}$ is the score of the top result in the current result set.
For each retrieve patent $p$ in the result set, the regression model produce a score $s_p$ which is used to reevaluate the score of $p$ to $w_p'$ such that $w_p' = w_p (s_p + 1)$. The final runs is obtained by re-sorting the list of results  and by applying a cutoff at rank 1000.

The regression model was trained using the normal training set and the additional validation set presented in the previous section. In order to limit the size of the training data and to avoid too many useless negative examples, we considered only 20 negative results per patent. The final run is also based on SVM regression, more precisely the WEKA implementation SMOreg using Pearson VII Universal Kernel (PUK) function. We did not evaluate other regression algorithms.

\section{Final Results}

\subsection{Main task}
Table \ref{evaluation} summarizes the automatic evaluation obtained for our final runs. We processed the entire list of queries (the XL set, corresponding to 10 000 patent topics) and, therefore, cover the smaller bundles (S and M, respectively 500 and 1000).  The list of relevant documents distinguishes two types of documents: {\sl relevant} documents (patents cited by the applicant and and with a A category) and {\sl very relevant} patents (all the other cited patents).

\begin{table}[ht]
\centering
\begin{tabular}{cc}

\begin{tabular}{l c c c}
\hline
Measures & S & M & XL  \\
\hline
MAP & 0.2714 & 0.2783 & 0.2802 \\
Prec. at 5 & 0.2780 & 0.2766 & 0.2768 \\
Prec. at 10 & 0.1768 & 0.1748 & 0.1776 \\
\hline
\end{tabular}
&
\begin{tabular}{l c c c}
\hline
Measures & S & M & XL  \\
\hline
MAP & 0.2832 & 0.2902 & 0.2836 \\
Prec. at 5 & 0.1856 & 0.1852 & 0.1878 \\
Prec. at 10 & 0.1156 & 0.1133 & 0.1177 \\
\hline
\end{tabular}
\\
\end{tabular}

\caption{\small Evaluation of official runs  for all relevant documents (left) and for highly relevant documents (right). }
\label{evaluation}
\end{table}

The exploitation of patent metadata and citation information clearly provides a significant improvement over a retrieval only based on textual data. By exploiting the same metadata as a patent examiner and combining them to robust text retrieval models via prior working sets and re-ranking, we created result sets closer to actual search reports. Overall, the exploitation of ECLA classes and of the patents cited in the descriptions provided the best improvements. In addition, the different regression models proved to be a very effective way of combining complementary indexing models. While many models exhibit relatively low individual results, they appear to be strongly specialized following discriminant criteria as the application language or the technical domain. Similarly a regression model appears an efficient technique for re-ranking a list of ranked results following heterogeneous features.

\subsection{Language tasks}

While the main task allows the participants to exploit all the available textual information, the language tasks limits the usable material to a single language. We applied for the language specific tasks a similar approach as for the main tasks, but with the following restrictions: 
\begin{itemize}
\item The patents cited in the description (which are used during the creation of initial working sets and in the final re-ranking) have been only considered for the main language of a patent topic ;
\item The lemma index corresponding to a different language was not used ; 
\item The phrase index was used only for English ;
\item The queries for concepts were limited to the concepts extracted in the text of a single language. 
\end{itemize}
We think that we have ensured, therefore, that for building the query and for the exploitation of patent metadata, we did not use any elements different from the task's language. 

\begin{table}[ht]
\centering

\begin{tabular}{l c c c c}
\hline
Measures & English & French & German & All  \\
\hline
MAP & 0.2358 &  0.1787 & 0.2092 & 0.2802  \\
Prec. at 5 & 0.2365 & 0.1855 & 0.2122 & 0.2768 \\
Prec. at 10 & 0.1575 & 0.1338 & 0.1467 & 0.1776 \\
\hline
\end{tabular}

\caption{\small Evaluation of official runs  for the three language tasks and the main multilingual task for the XL bundle. }
\label{languages}
\end{table}

Table \ref{languages} presents the evaluation of the three language tasks compared to the multilingual one. We can observe that the exploitation of English language clearly leads to better results as compared to the other languages. This result is not surprising because almost 70\% of the textual material, in particular the descriptions, are in English.
However, the combination of all languages was effective and has provided the best performance.

\section{Hardware and Processing Time}

We use 3 machines with 2.0GHz Core 2 Duo processor, 2 GB SDRAM (3 headless Mac Mini) and a Laptop with a similar configuration and 4 GB RAM. In addition 2 TB of storage have been used for the collection files, indexes and all the intermediary processed documents. The four machines ran under 64bits Mac OS 10.5.6. 
Here are some indications about  the processing time:
\begin{itemize}
\item The compilation of the terminological database took 28 hours on one machine after a pre-processing of 135 hours for the three language Wikipedia XML files.
\item Tokenization and POS tagging took 55 hours-machine. Phrase extraction was the heaviest task and took a total of approx. 28 days-machine.
\item Controlled concept indexing took approx. 22 hours-machine.
\item Training a regression model took between 10 minutes (least median squared linear regression) to 150 minutes (Multilayer Perceptron) per model. Aggregation of results and post-ranking took approx. 90 minutes.
\item Producing the final runs required 5 complete days of processing using entirely the 4 machines, i.e. approx. 20 days-machine. 
\end{itemize}

The total processing time for a topic patent was, therefore, approx. 43 secondes. This is, of course, quite long for an online processing. However, given the challenge of processing the whole collection of patent documents, we did not address at all the issue of processing time and many possibilities for runtime improvement and optimization exist.

\section{Future Work}

We have tried in the present work to create a theoritically sound framework that could be generalized to non patent and mixed collection of technical and scientific documents. 
More complementary retrieval models and more languages can easily be added to the current architecture of PATATRAS. 
If some metadata are specific to patent information, many of them find their counterpart in non patent articles. Fine-grained and theoretically well founded taxonomies such as MeSH can be used instead of patent classifications. 

Although our system topped the evaluation for all tasks and all size of patent topics, we do not consider that any aspects of the present system are finalized. First, we have chosen not to focus on query formulation and we have simply selected the whole patent topic representation as query. Using more sophisticated query formulation such as parsimonious language models and an improved analysis of the patent structures could provide interesting improvements and make query expansion and pseudo-relevant feedback effective. Regarding the machine learning approach, a larger set of features combined with feature selection methods would need to be used.
The initial creation of working sets should also be expressed as a machine learning problem so that the selection of interesting patents could be realized on patent topic by patent topic basis. 
The exploitation of citation text is also currently disappointing because of the limited number of cited EP documents. While this problem could be solved by exploiting patent family information, it might also be well tackled for non patent litterature in repository services as HAL or CiteSeer by using automatic bibliographical extraction techniques such as the one presented in \cite{grobid}.

Finally, the terminological database suffers from the low quality of data of Wikipedia which introduces a considerable noise and makes the merging of concepts frequently erroneous. In future, we plan to experiment more controlled terminological multi-domain and multilingual large scale resources. 

\bibliographystyle{plain}

\end{document}